\documentclass[10pt,twocolumn,letterpaper]{article}

\usepackage{iccv}
\usepackage{times}
\usepackage{epsfig}
\usepackage{graphicx}
\usepackage{amsmath}
\usepackage{amssymb}

\usepackage{booktabs}
\usepackage[makeroom]{cancel}
\usepackage[table,xcdraw,dvipsnames]{xcolor}
\usepackage{adjustbox}

\usepackage[breaklinks=true,bookmarks=false]{hyperref}

\iccvfinalcopy % *** Uncomment this line for the final submission

\def\iccvPaperID{****} % *** Enter the ICCV Paper ID here

\ificcvfinal\fi

\begin{document}

%%%%%%%%% TITLE
% \title{USD: Enhancing Open World Object Detection via \underline{D}ecoupled Objectness and \\ 
% \underline{L}arge \underline{V}ision \underline{M}odels}
\title{USD: \underline{U}nknown \underline{S}ensitive \underline{D}etector Empowered by Decoupled Objectness and  
Segment Anything Model}

\author{Yulin He\thanks{Equal contribution.}    ~~~~~~~
Wei Chen\footnotemark[\value{footnote}] \thanks{Corresponding author.}  ~~~~~~~
Yusong Tan ~~~~~~~ Siqi Wang\\
 National University of Defense Technology \\
{\tt\small \{heyulin, chenwei, ystan, wangsiqi10c\}@nudt.edu.cn}}

\maketitle
\ificcvfinal\thispagestyle{empty}\fi

%%%%%%%%% ABSTRACT
\begin{abstract}
    Open World Object Detection (OWOD) is a novel and challenging computer vision task that enables object detection with the ability to detect unknown objects.
    Existing methods typically estimate the object likelihood with an additional objectness branch, but ignore the conflict in learning objectness and classification boundaries, which oppose each other on the semantic manifold and training objective. 
    To address this issue, we propose a simple yet effective learning strategy, namely Decoupled Objectness Learning (DOL), which divides the learning of these two boundaries into suitable decoder layers.
    Moreover, detecting unknown objects comprehensively requires a large amount of annotations, but labeling all unknown objects is both difficult and expensive.
    Therefore, we propose to take advantage of the recent Large Vision Model (LVM), specifically the Segment Anything Model (SAM), to enhance the detection of unknown objects.
    Nevertheless, the output results of SAM contain noise, including backgrounds and fragments, so we introduce an Auxiliary Supervision Framework (ASF) that uses a pseudo-labeling and a soft-weighting strategies to alleviate the negative impact of noise.
    Extensive experiments on popular benchmarks, including Pascal VOC and MS COCO, demonstrate the effectiveness of our approach.
    Our proposed Unknown Sensitive Detector (USD) outperforms the recent state-of-the-art methods in terms of Unknown Recall, achieving significant improvements of 14.3\%, 15.5\%, and 8.9\% on the M-OWODB, and 27.1\%, 29.1\%, and 25.1\% on the S-OWODB.
\end{abstract}

%%%%%%%%% BODY TEXT
\section{Introduction}

Object detection (OD) is a critical computer vision task with significant implications in various fields such as autonomous driving~\cite{ma2022rethinking,li2022coda}, smart healthcare~\cite{elakkiya2022imaging,karri2022multi}, and intelligent robots~\cite{andronie2023big, jiang2022lightweight}. 
However, traditional OD methods are predominantly designed and evaluated in closed-set environments, which greatly restricts their applicability in real-world scenarios.
Closed-set OD methods suffer from several limitations, such as their tendency to misclassify unknown objects as known, their inability to detect unknown objects, and their incapacity to learn incrementally as the number of images and labels increases. 
To address these challenges, a more demanding task called Open-World Object Detection (OWOD) has emerged.
In OWOD, models are not only expected to identify objects of known categories defined in the dataset but also to identify  potential unknown objects, i.e., objects that have not seen before.
The detected unknown objects are then presented to an oracle, typically a human annotator, who labels some objects of interest.
The newly labeled data is subsequently incorporated into an incremental learning paradigm, allowing the model to quickly learn and adapt to new classes while mitigating catastrophic forgetting of previously learned classes.

Undoubtedly, the detection of unknown objects holds immense significance in the context of the OWOD task. 
However, the absence of labels for unknown objects often leads to misclassify them as background during the training phase, resulting in a low recall for detecting unknown objects.
To overcome this issue, existing OWOD methods have incorporated an additional objectness branch to differentiate between unknown objects and backgrounds. 
For instance, in ~\cite{towards, du2022vos, liang2023unknown}, an energy-based classifier has been adopted to distinguish known and unknown objects based on their energy scores.
In ~\cite{Owdetr,UCOWOD,CAT,revisiting}, a pseudo-labelling strategy has been employed, selecting top-k confident regions as pseudo-labels to supervise the training of the objectness branch.
In a recent study by PROB~\cite{prob}, a two-stage approach is proposed, where an objectness probabilistic model is developed in the first stage to differentiate objects from backgrounds, then a classification model is trained in the second stage to distinguish specific classes.
Notably, the probabilistic model is trained without negative samples, avoiding confusion between unknown objects and backgrounds.
However, these methods do not adequately address the conflict arising from learning objectness and classification boundaries.
Moreover, the availability of annotations plays a crucial role in the performance of unknown object detection, as only when objects are sufficiently annotated can the detector exhibit comprehensive detection of unknown objects.

The challenge of learning objectness and classification boundaries arises due to the inherent conflict between semantic manifolds and training objectives.
The objectness boundary requires low-level semantics to comprehensively detect objects, whereas the classification boundary relies on high-level semantics to accurately classify objects into specific categories.
Furthermore, the objectness boundary aims to minimize the distance between known objects, while the classification boundary aims to maximize the distance between different object classes.
To tackle these two main conflicts, we propose a simple yet effective learning strategy, termed Decoupled Objectness Learning (DOL), which addresses this conflict by segregating the learning of these two boundaries into distinct decoder layers.
Specifically, we assign the learning of objectness to the first decoder layer, which operates as a category-agnostic classifier. 
Subsequently, the remaining decoder layers refine localization and perform category-specific classification.
By decoupling the learning of objectness and classification boundaries, the traits of objectness and classification are fully released.

Another challenge in developing OWOD is the limited annotation of unknown objects.
It is evident that labeling more unknown objects will expand the detector's capability to detect a wider range of objects. 
However, annotating unknown objects is challenging due to the difficulty in distinguishing them from the background.
To overcome this challenge, we propose leveraging the zero-shot and open-world capabilities of Large Visual Models (LVMs)~\cite{clip,glip,sam,seea} without requiring human annotations.
Specifically, we harness the recently proposed Segment Anything Model (SAM)~\cite{sam}, which can achieve outstanding performance of class-agnostic localization with the help of large-scale data.
However, SAM's output may include regions that contain backgrounds and fragments, which can be detrimental to unknown object detection. 
To address this issue, we introduce an Auxiliary Supervision Framework (ASF), which employs the objectness score and the Intersection-over-Union (IoU) between prediction boxes and SAM's output boxes as reliable indicators to filter out noise.
Subsequently, we utilize these filtered unknown pseudo-labels as auxiliary supervision to facilitate the learning process of unknown object detection, leading to a more comprehensive detection of unknown objects.

In general, we summarize our main contributions as follows:

\begin{itemize}
  \item We introduce a simple yet effective learning strategy to address the inherent conflict between learning objectness and classification boundaries, enhancing the detection of known and unknown objects without incurring additional costs.
  \item We are the first to propose harnessing the zero-shot and open-world capabilities of SAM to solve the annotation dilemma of unknown objects, achieving a more comprehensive detection for unknown objects.
  \item We design a generic framework that effectively mitigates the negative effects of noise in the SAM output, significantly improving the performance of both known and unknown object detection.
  \item Through rigorous evaluation on two widely-utilized data splits of the Pascal VOC and MS COCO datasets, our proposed method achieves a remarkable improvement in unknown recall, establishing a new state-of-the-art in the OWOD task.
\end{itemize}

\section{Related Works}

\subsection{Open-World Object Detection}
Open-world object detection aims to address the limitations of close-set and static-learning settings in traditional object detection, recently proposed by ORE~\cite{towards}, has gained substantial attention.
It can be split into two sub-tasks, i.e., known and unknown object detection, and incremental object detection~\cite{shmelkov2017incremental,yang2022multi}.
The latter has been extensively researched as a typical incremental learning task~\cite{zhou2023deep}, while the former is still in its infancy stage as a task specific to OWOD.
To detect known and unknown objects, two classification boundaries need to be learned: one that distinguishes objects from backgrounds (i.e., objectness boundary), and one that distinguishes objects into specific classes (i.e., classification boundary).
The learning of the second boundary is the goal of open-set object detection task and has been studied in a number of works~\cite{zheng2022towards,overlooked,expanding,uncertainty}, while the learning of the first boundary is challenging since we cannot define all categories of objects, and the objectness is a subjective concept that is easily confused.
Class-agnostic object detection~\cite{MViTs,zhao2022exploiting,kim2022learning,saito2022learning} is the most related task, which considers all objects as the objectness category and the others as the background category.
However, it requires a large amount of data to enable the detection of more objects~\cite{MViTs,kim2022learning}.
In OWOD, the crucial challenge is to detect unknown objects comprehensively while maintaining high accuracy in detecting known objects.
ORE~\cite{towards} proposes an energy based model to identify known and unknown objects.
2B-OCD~\cite{wu2022two} proposes an additional IoU-based location branch for more accurate estimation of objectness.
UC-OWOD~\cite{UCOWOD} proposes to exploit the pseudo-labeling paradigm, which selects the filtered top-k potential unknown regions to train objectness.
Recently, Transformer-based methods have shown great potential in learning objectness, OW-DETR~\cite{Owdetr} first adapted the deformable DETR~\cite{ddetr} model for OWOD.
It exploits a model-driven pseudo-labeling scheme to supervise unknown objectness learning.
Following this, CAT~\cite{CAT} utilized the selective search algorithm~\cite{selective} to generate class-agnostic auxiliary proposals, resulting in high-recall pseudo-labels.
On the other hand, PROB~\cite{prob} proposed an anomaly detection based approach, which doesn't require negative examples and thus avoids the confusion between background and unknown objects.
Different from the above methods, this paper reveals the conflict problem that arises in learning objectness and classification boundaries, and proposes to benefit from SAM to alleviate the annotation dilemma in OWOD.

\subsection{Large Visual Model}
The success of Large Language Models (LLMs)~\cite{bert,brown2020language} such as ChatGPT~\cite{chatgpt} has shown the superiority and significance of large-scale models.
Consequently, the research community has devoted considerable attention to Large Vision Models (LVMs). 
CLIP~\cite{clip}, for instance, has emerged as a pivotal advancement that bridges the gap between the domains of Natural Language Processing (NLP) and Computer Vision (CV) by constructing image-text pairs, providing a novel solution to model the open-world vision space with impressive zero-shot and generalization abilities.
In addition to the classification task, LVMs have been extended to object detection~\cite{MViTs,gu2021open,kuo2022f} and segmentation~\cite{sam,seea}, achieving remarkable performance improvements in open-world settings.
Recently, Segment Anything Model (SAM)~\cite{sam} has demonstrated the powerful zero-shot segmentation capability in the open environment and can segment anything using appropriate prompts, such as points and boxes.
Notably, in the ``everything'' mode of SAM, it can utilize a $n \times n$ grid of sampling points across the entire image, with each point generating a mask to segment potential objects, which can provide high-recall auxiliary proposals to enable OWOD methods to detect more objects.
However, the results of SAM may contain backgrounds and fragments that are detrimental to unknown object detection.
This paper addresses this issue by exploiting a pseudo-labeling and a soft-weighting strategies, alleviating the negative impact of noise, while still benefiting from the high-recall unknown proposals of SAM.

\begin{figure*}[t]
    \centerline{\includegraphics[width=1.0\linewidth]{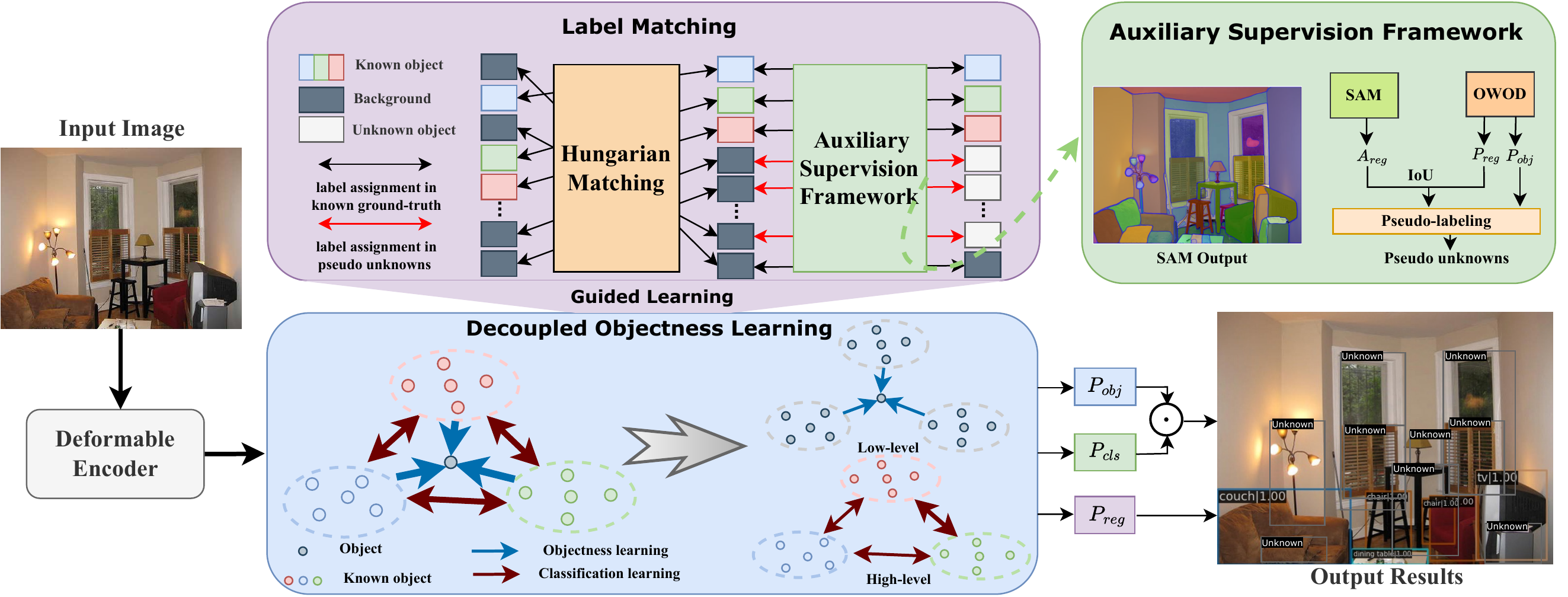}}
    \caption{
      Our proposed method is based on DDETR~\cite{ddetr}, which has a deformable encoder and a deformable decoder.
      To adapt DDETR into the open-world setting, a decoupled objectness learning and an auxiliary supervision framework are proposed.
      The decoupled objectness learning strategy separates the learning of objectness and classification into different decoder layers.
      Specifically, the first decoder layer models the objectness score, while the remaining decoder layers predict the classification and regression results.
      The auxiliary supervision framework combines the pseudo-labeling scheme with LVMs, which alleviates the annotation dilemma in OWOD.
    } \label{fig:overview}
\end{figure*}

\section{Method}
We propose USD, which adapts DDETR~\cite{ddetr} model for open-world object detection, with our proposed decoupled objectness learning strategy and the auxiliary supervision framework.
In Sec.~\ref{sec:preliminary}, we provide an introduction to the problem of OWOD and briefly outline our baseline method, PROB~\cite{prob}.
In Sec.~\ref{sec:DOD}, we delve into the inherent conflict between learning objectness and classification boundaries and present our solution to alleviate this conflict.
Furthermore, in Sec.~\ref{sec:ASF}, we show the noise problem in SAM output and describe our approach to mitigate its negative impact.
Fig.~\ref{fig:overview} illustrates the overview of USD, a transformer-based open-world detector, with strong perceptual capability towards unknown objects.

\subsection{Preliminary} \label{sec:preliminary}
\textbf{Problem definition:} 
In standard object detection, a set of known classes $K = \{1, 2, 3,..., C \} \in N^{+}$ is given, along with the corresponding training set $D^t = \{X^t, Y^t \}$ consisting of images $X$ and their corresponding labels $Y$.
Each image $X_i$ contains multiple class-specific instances, with one instance labeled as class $l_i \in K$ and bounding box $b^t_i = \{x^t, y^t, w^t, h^t \}$, where $(x^t, y^t)$ represents the center coordinate, $w^t$ and $h^t$ represent the width and height, respectively.
In addition to the known classes, the images $X$ may also contain unknown objects with classes $U = \{C+1, C+2, ... \}$.

OWOD methods aim to detect potential unknown objects and feed them to an oracle to annotate new classes for continually learning.
Building upon the pioneering work of \cite{towards}, OWOD methods typically involve a set of sequential tasks $T = \{T_1, T_2, T_3, T_4\}$.
A model $M^t$ is trained at time $t$ ($t \in T$), accessing the training data with known annotations $Y^t$, the previous training model $M^{t-1}$, and a small subset of the previous training data $D_s^{t-1}$, where $D_s^{t-1}  \subset  D^{t-1}$ and $ | D_s^{t-1} | <<  | D^{t-1}|$.
In Task $T_{t+1}$, the known classes are continuously increased $K^{t+1} = K^{t} \cup \{C+1, C+2, ... C+n\}$ and unknown classes $U^{t+1} = \{C+n+1, ...\}$, where $n$ is the number of newly labeled classes.
The model will be updated to $M^{t+1}$ without learning from scratch, and can detect not only the previous classes $K^{t}$ but also the recent labeled classes $\{C+1, C+2, ..., C+n \}$, along with the unknown objects.
This process supports multi-round human-AI interactions and is a lifelong learning paradigm, which is more practical for open-world environments.

\textbf{Baseline:} 
The baseline of our method is PROB~\cite{prob}, a recent proposed state-of-the-art OWOD method.
PROB separates the unknown object detection into objectness prediction $p(o)$ and object class prediction $p(c \mid o)$, which can be seen as a two-stage probabilistic model~\cite{zhou2021probabilistic}.
The inference scheme can be formalized as follows:
\begin{equation}
    p(c \mid q)=p(c \mid o, q) \cdot p(o \mid q).
\end{equation}Therefore, there are two classification boundaries needed to learn:
\begin{equation}
    p(c \mid q)= \mathcal{F}_{cls}(q) \cdot \mathcal{F}_{obj}(q),
\end{equation}i.e., the objectness boundary $\mathcal{F}_{obj}$ and the classification boundary $\mathcal{F}_{cls}$.
The former distinguishes between object and background, while the latter classify specific classes.
To avoid the confusion problem in judging objects and backgrounds at the first boundary, PROB exploits the idea of anomaly detection, which minimizes the distance of known objects without requiring negative samples, formalized as:
\begin{equation}
    \mathcal{L}_{obj}=\sum_{i \in Z} d_M\left({q}_i\right)^2,
    \label{eq:obj}
\end{equation}where $d_M$ is the Mahalanobis distance, and $Z$ is a list of indices of known queries.

\begin{figure}[t]
    \centerline{\includegraphics[width=1.0\linewidth]{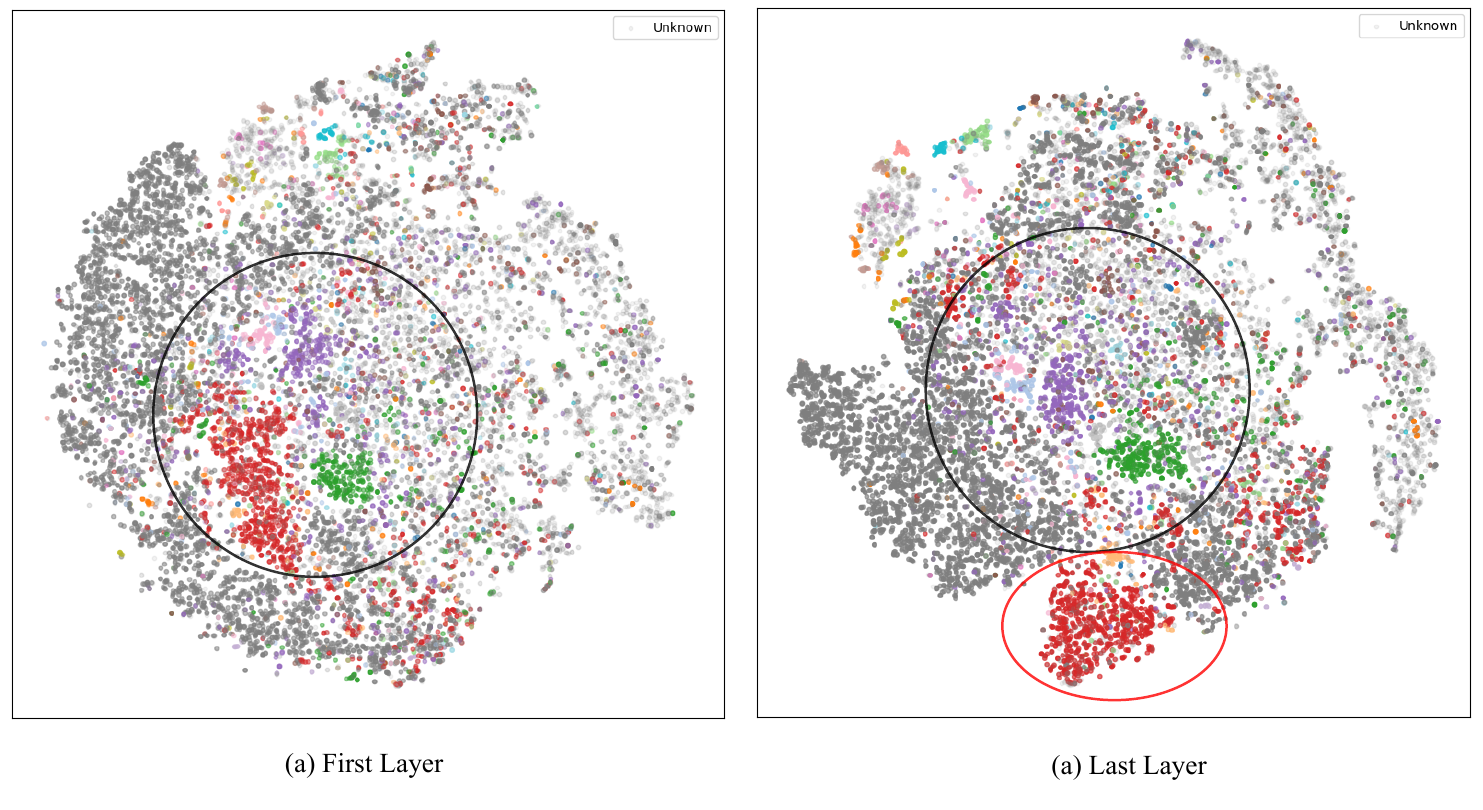}}
    \caption{The t-SNE analysis in latent features of the first and the last decoder layers.
    We conduct this experiment by randomly sampling 1k images from the COCO validation set with the pretrained PROB model.
    Colored dots represent known objects with specific classes and grey dots are unknown objects. 
    } \label{fig:tsne}
\end{figure}

\subsection{Decoupled Objectness Learning} \label{sec:DOD}

In existing methods such as OW-DETR~\cite{Owdetr}, PROB~\cite{prob}, and CAT~\cite{CAT}, objectness, classification, and regression tasks are learned simultaneously in each decoder layer.
However, these methods fail to address the conflict that arises between learning the objectness and classification boundaries.
The objectness boundary aims to minimize the distance between known objects, which helps in distinguishing unknown objects that are often represented as outliers. 
On the other hand, the classification boundary aims to maximize the distance between objects belonging to different classes, enabling the discrimination of specific categories.
Consequently, the training objectives of these two boundaries conflict with each other.
Moreover, the suitable semantic manifolds for these two boundaries also differ. 
To investigate this issue, we present t-SNE plots~\cite{t-SNE} of the latent features obtained from the first and last decoder layers in Fig.~\ref{fig:tsne}. 
At the first layer of the decoder, the known object features are concentrated in the central region of the manifold, aligning with the objectives of objectness learning.
However, at the last layer of the decoder, the feature points represented in red shift to the boundary region of the manifold, which contradicts the objective of objectness learning.
Nonetheless, the distance between objects of different classes is increased, thereby fulfilling the objectives of classification learning.

To handle the conflict problem, we introduce a subtle approach to separately learn the objectness and classification boundaries by utilizing different decoder layers. 
As shown in the DOL part of Fig.~\ref{fig:overview}, we introduce a brief overview of our solution.
Specifically, we assign the first decoder layer to solely focus on the objectness task, with the aim of localizing as many objects as possible within the limited number of queries. 
Since the queries are dynamic and trainable, they already contain localization information, and thus, we do not add the regression loss at the first decoder layer. 
The subsequent decoder layers are dedicated to the classification and regression tasks, progressively refining object predictions.
During inference, the final prediction result is determined by the product of the objectness score from the first layer and the classification score from the last layer, formalized as: 
\begin{equation}
    p(c \mid q)= \mathcal{F}^{last}_{cls}(q) \cdot \mathcal{F}^{first}_{obj}(q).
\end{equation}
By decoupling the training objectives and semantic manifolds of the two tasks, our method enables them to concentrate on their respective duties, effectively alleviating the conflict.
This approach mimics the human perceptual process of first perceiving potential object locations globally and then focusing on specific object categories.

\begin{figure}[t]
    \centerline{\includegraphics[width=1.0\linewidth]{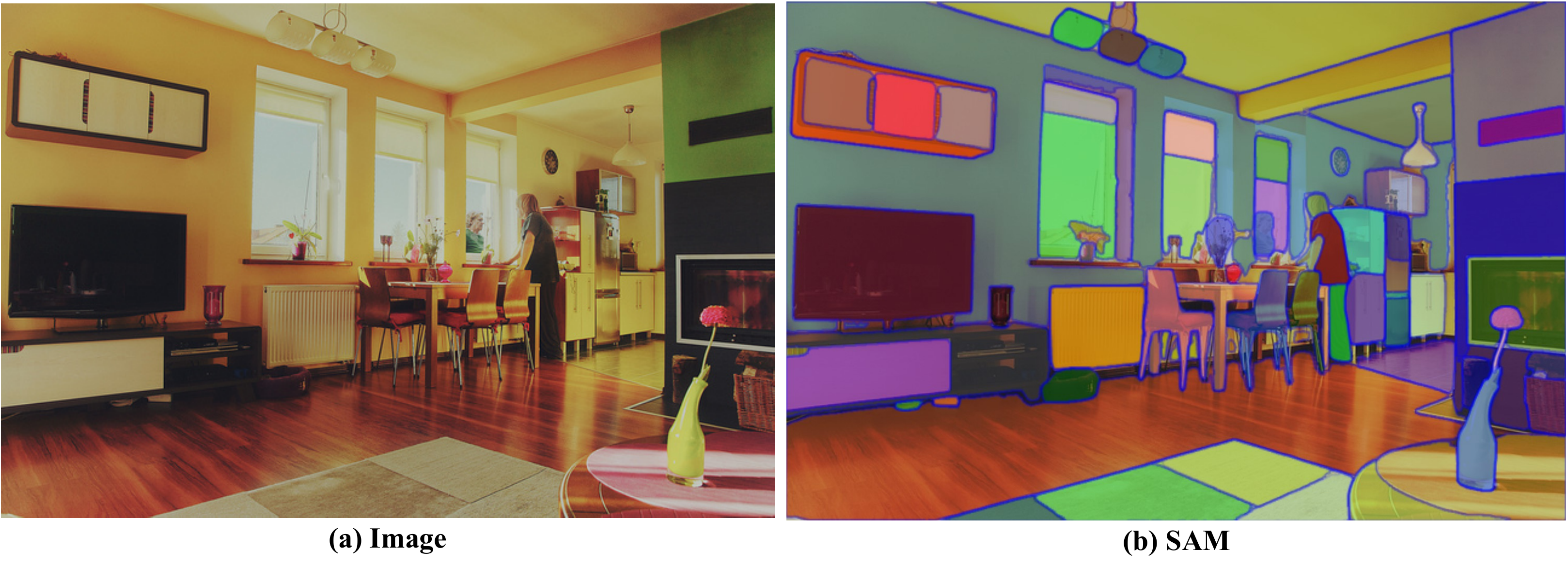}}
    \caption{Segmentation results in ``everything'' mode of SAM. 
    } \label{fig:sam}
\end{figure}

\subsection{Auxiliary Supervision Framework}  \label{sec:ASF}

SAM segments potential objects from the entire image by sampling a grid of prompt points, each of which predicts a mask that may contain a valid object.
Then, we generate auxiliary boxes $B_{aux}$ from the minimum external rectangle of each mask.
However, as shown in Figure~\ref{fig:sam}, the segment results may contain noise such as background (e.g., walls, ceilings, and floors) and fragments (e.g., head, trunk, and leg in a person).
In the broad definition of objectness, some parts of an object can also be regarded as objects, such as the face, hand, and feet of a person, or the eye, nose, and mouth of a face, but at different levels, i.e., ``whole'', ``part'', and ``subpart''.
We argue that detecting objects at all levels is not feasible, the level at which the model detects should be determined by the annotation level of the dataset.
For instance, in the current OWOD datasets like MS COCO~\cite{mscoco}, objects are all categorized at the ``whole'' level, which means that some ``part'' or ``subpart'' objects may not be supposed to recognized as objects.

To address this issue, we propose an auxiliary supervision framework (ASF) to alleviate the negative impact of noise in SAM's output.
We first remove auxiliary boxes that have a large overlap with the currently known target boxes, getting potential unknown boxes.
Then, we define a matching cost to filter out the noise in the results of SAM.
The matching cost involves two main variables, the objectness score $P_{obj}$ and the $IoU$ between prediction boxes $P_{box}$ and $B_{aux}$.
$P_{obj}$ is designed to identify objects in OWOD, which helps in removing some backgrounds.
$IoU$ between $P_{box}$ and $B_{aux}$ is used to filter out some fragments, as the predicted boxes can reflect the level of the unknown object in some extent.
These variables are combined using a geometric mean to form the matching cost function:
\begin{equation}
    C = P_{obj} ^ \alpha \cdot IoU (P_{box}, B_{aux}) ^ {1- \alpha}
    \label{eq:score}
\end{equation}where $\alpha$ is a hyperparameter to control the contribution. 
Afterwards, we sort potential unknown boxes using the matching cost and filter out the noise with a threshold, getting the pseudo unknown objects $B_{pse}$ to aid models for unknown learning. 
Then, we need to assign labels for predictions, a Hungarian algorithm is performed to form a one-to-one matching on predictions and ground-truth boxes, where the matched indexes $I_m$ correspond to known objects and the unmatched indexes $I_{um}$ correspond to backgrounds.
However, the assigned backgrounds contain potential unknown objects.
Therefore, we assign some of the backgrounds with pseudo unknown objects as auxiliary supervision in unmatched indexes, promoting the learning of unknown objects.
Based on the matching cost in Equation~\ref{eq:score}, another hungarian algorithm is performed to ensure the one-to-one matching in unmatched predictions and pseudo unknown objects, where $I_{pm}$ are pseudo matched indexes.

Therefore, ASF has an additional auxiliary loss, which consists of auxiliary objectness loss and auxiliary regression loss.
The auxiliary objectness loss can be formulated as :
\begin{equation}
    \mathcal{L}_{aux}^{pse}=\sum_{i \in I_{pm}} W_i \cdot d_M\left({q}_i\right)^2,
    \label{eq:pseobj}
\end{equation}where $ W_i = exp(-T \cdot d_M(q_i)^2)$ represents the objectness likelihood expressed as an energy score and the temperature $T$ is set to 1.3, the same as ~\cite{prob}.
Compared with Eq~\ref{eq:obj}, Eq~\ref{eq:pseobj} has soft weights ranging from 0 to 1, which focus on the learning of confident pseudo unknown labels, alleviating the impact of noise during training.
Similar to the pseudo objectness loss, we also define an auxiliary regression loss with soft weights.
\begin{equation}
    \mathcal{L}_{aux}^{pse}=\sum_{i \in I_{pm}} W_i \cdot d_{L1}\left(\mathcal{F}_{reg}({q}_i), B_{pse}\right), \\
    \label{eq:psel1}
\end{equation}where $d_{L1}$ is the MAE (Mean Absolute Error) distance.

The total loss in our USD is as follows:
\begin{equation}
    \mathcal{L}= \underbrace {\mathcal{L}_{cls} + \mathcal{L}_{reg}}_{\text{OD Loss}} + \underbrace{\mathcal{L}_{obj}}_{\text{Obj. Loss}} + \underbrace {\lambda_{obj}^{pse} \mathcal{L}_{obj}^{pse} +  \lambda_{reg}^{pse} \mathcal{L}_{reg}^{aux}}_{\text{Aux. Loss}},
\label{eq:loss}
\end{equation}consisting of three parts, i.e., standard OD loss, objectness loss, and auxiliary loss in ASF.
For the OD loss and objectness loss, we adopt the same coefficient values as in PROB \cite{prob} for consistency and comparability.
To balance the importance of the pseudo loss with the other losses, we set $\lambda_{reg}^{pse}$ to be the same as in $\mathcal{L}_{reg}$, and the coefficient $\lambda_{obj}^{pse}$ is set to be one-tenth of the coefficient in $\mathcal{L}_{obj}$.

\section{Experiments} 

\subsection{Dataset}
\label{sec:dataset}
The experiments are conducted on two mainstream data splits of MS COCO and Pascal VOC, introduced by ~\cite{towards} and ~\cite{Owdetr}, which can be considered as a ``superclass-mixed OWOD benchmark'' (M-OWODB) and a ``superclass-separated OWOD benchmark'' (S-OWODB), respectively.
In M-OWODB, images from MS COCO~\cite{mscoco} and Pascal VOC~\cite{everingham2010pascal} are splited into four sets of non-overlapping tasks $T = \{T_1, T_2, T_3, T_4\}$.
When reaching to $T_t$, an additional 20 classes are introduced, and all classes seen before $\left\{T_\lambda: \lambda \leq t\right\}$ should be detected.
In S-OWODB, a clear super-categories separation is proposed, which groups object categories with the close semantic into a task, again divided into four tasks.
Clearly, S-OWODB is more challenging than M-OWODB, as S-OWODB is superclass-separated and the semantic of unknown objects is harder to expand.
See more details in Appendix \ref{sec:sup:data}.

\subsection{Evaluation Metric}
Following the most commonly used validation metric in OWOD~\cite{towards,Owdetr,CAT,prob}, we use mean average precision (mAP) to evaluate the known object detection.
To better understand the quality of continual learning, mAP is partitioned into previously and newly introduced object classes.
For unknown object detection, as we do not have all the annotations of unknown objects, we use unknown object recall (U-recall) as the main metric instead of mAP.
To study the confusion of unknown object detection, Wilderness Impact (WI) and Absolute Open-Set Error (A-OSE) are used, the results of which are shown in Appendix~\ref{sec:sup:openset}.

\subsection{Implementation Details}
Following the recent transformer-based OWOD methods~\cite{Owdetr,prob,CAT}, we use DDETR~\cite{ddetr} model with a DINO-pretrained ResNet-50 FPN backbone~\cite{dino,resnet,fpn} as the default detector.
After extracting multi-scale features, the trainable queries ($N_{query}=100$) are input to the deformable decoder, getting query embeddings ($D_{query}=256$).
Each of queries predicts a box and a score, top-100 high scoring detections per image are used for evaluation during inference.
The geometric parameter $\alpha=0.5$ and the pseudo-labeling threshold is set to 0.7.
Additional details are provided in Appendix~\ref{sec:sup:detail}.

\begin{table*}[t]
   \centering
   \caption{\textbf{State-of-the-art comparison for OWOD on both M-OWODB (top) and S-OWODB (bottom).} 
   The comparison is presented in terms of two crucial evaluation metrics: unknown class recall (U-Recall) and known class mAP at an IoU threshold of 0.5 (mAP@0.5) for previously, currently, and all known objects.
   Note that U-Recall is not computed in Task 4, as all 80 classes are known in this case. }
   \setlength{\tabcolsep}{2pt}
   \adjustbox{width=\textwidth}
   {
   \begin{tabular}{@{}l|cc|cccc|cccc|ccc@{}}
   \toprule
    \textbf{Task IDs} ($\rightarrow$)& \multicolumn{2}{c|}{\textbf{Task 1}} & \multicolumn{4}{c|}{\textbf{Task 2}} & \multicolumn{4}{c|}{\textbf{Task 3}} & \multicolumn{3}{c}{\textbf{Task 4}} \\ \midrule
    
   & \cellcolor[HTML]{FFFFED}{U-Recall} & \multicolumn{1}{c|}{\cellcolor[HTML]{EDF6FF}{mAP ($\uparrow$)}} & \cellcolor[HTML]{FFFFED}{U-Recall} & \multicolumn{3}{c|}{\cellcolor[HTML]{EDF6FF}{mAP ($\uparrow$)}} & \cellcolor[HTML]{FFFFED}{U-Recall} & \multicolumn{3}{c|}{\cellcolor[HTML]{EDF6FF}{mAP ($\uparrow$)}} & \multicolumn{3}{c}{\cellcolor[HTML]{EDF6FF}{mAP ($\uparrow$)}}  \\

    & \cellcolor[HTML]{FFFFED}($\uparrow$) & \begin{tabular}[c]{@{}c}Current \\ known\end{tabular} & \cellcolor[HTML]{FFFFED}($\uparrow$) & \begin{tabular}[c]{@{}c@{}}Previously\\  known\end{tabular} & \begin{tabular}[c]{@{}c@{}}Current \\ known\end{tabular} & Both & \cellcolor[HTML]{FFFFED}($\uparrow$) & \begin{tabular}[c]{@{}c@{}}Previously \\ known\end{tabular} & \begin{tabular}[c]{@{}c@{}}Current \\ known\end{tabular} & Both & \begin{tabular}[c]{@{}c@{}}Previously \\ known\end{tabular} & \begin{tabular}[c]{@{}c@{}}Current \\ known\end{tabular} & Both \\ \midrule

   ORE\texttt{-EBUI}~\cite{towards} & \cellcolor[HTML]{FFFFED} 4.9  & 56.0 & \cellcolor[HTML]{FFFFED}2.9 & 52.7 & 26.0 & 39.4  & \cellcolor[HTML]{FFFFED}3.9 & 38.2 & 12.7 & 29.7 & 29.6 & 12.4 & 25.3 \\ 
   
   UC-OWOD~\cite{UCOWOD} & \cellcolor[HTML]{FFFFED} 2.4  & 50.7 & \cellcolor[HTML]{FFFFED} 3.4 & 33.1 & 30.5 & 31.8  & \cellcolor[HTML]{FFFFED} 8.7 & 28.8 & 16.3 & 24.6 & 25.6 & 15.9 & 23.2 \\ 
   
   OCPL~\cite{OCPL} & \cellcolor[HTML]{FFFFED} 8.26  & 56.6 & \cellcolor[HTML]{FFFFED} 7.65 & 50.6 & 27.5 & 39.1  & \cellcolor[HTML]{FFFFED} 11.9 & 38.7 & 14.7 & 30.7 & 30.7 & 14.4 & 26.7 \\ 
   
   2B-OCD~\cite{wu2022two} & \cellcolor[HTML]{FFFFED} 12.1  & 56.4 & \cellcolor[HTML]{FFFFED} 9.4 & 51.6 & 25.3 & 38.5  & \cellcolor[HTML]{FFFFED} 11.6 & 37.2 & 13.2 & 29.2 & 30.0 & 13.3 & 25.8 \\ 
   
   OW-DETR~\cite{Owdetr} & \cellcolor[HTML]{FFFFED}7.5  & 59.2 & \cellcolor[HTML]{FFFFED}6.2 & 53.6 & \textbf{33.5} & 42.9 & \cellcolor[HTML]{FFFFED}5.7 & 38.3 & 15.8 & 30.8 & 31.4 & 17.1 & 27.8 \\

   CAT \texttt{-cddw}~\cite{CAT}  & \cellcolor[HTML]{FFFFED} 19.8  & 59.3 & \cellcolor[HTML]{FFFFED} 16.8 & 53.0 & 29.4 & 41.3  & \cellcolor[HTML]{FFFFED} 21.8 & 38.1 & 15.0 & 30.5 & 30.6 & 14.0 & 26.8 \\ 
   
   CAT$^ \dag$~\cite{CAT}  & \cellcolor[HTML]{FFFFED} 21.8  & 59.5& \cellcolor[HTML]{FFFFED} 19.2 & 54.6 & 32.6 & 43.6  & \cellcolor[HTML]{FFFFED} 24.4 & 42.3 & 18.9 & 34.5 & 34.4 & 16.6 & 29.9 \\ 

   PROB~\cite{prob}  & \cellcolor[HTML]{FFFFED} 19.4 & 59.5 & \cellcolor[HTML]{FFFFED} 17.4 & 55.7 & 32.2 & 44.0  & \cellcolor[HTML]{FFFFED} 19.6 & \textbf{43.0} & \textbf{22.2} & 36.0 & \textbf{35.7} & \textbf{18.9} & \textbf{31.5} \\

   \textbf{Ours: USD} \texttt{-ASF}  & \cellcolor[HTML]{FFFFED} 21.6  & \textbf{59.9 }& \cellcolor[HTML]{FFFFED} 19.7 & \textbf{56.6} & 32.5 & \textbf{44.6}  & \cellcolor[HTML]{FFFFED} 23.5 & \textbf{43.5} & 21.9 & \textbf{36.3} & 35.4 & \textbf{18.9} & 31.3 \\ 
   
   \textbf{Ours: USD $^ \ddag$}  & \cellcolor[HTML]{FFFFED} \textbf{36.1}  & 58.4& \cellcolor[HTML]{FFFFED} \textbf{34.7} & 54.3 & 31.4 & 42.7  & \cellcolor[HTML]{FFFFED} \textbf{33.3} & 41.5 & 20.5 & 34.5 & 33.4 & 16.6 & 29.2 \\ 
   
   \midrule
   \midrule 
   
   ORE\texttt{-EBUI}~\cite{towards} & \cellcolor[HTML]{FFFFED}1.5  & 61.4 & \cellcolor[HTML]{FFFFED}3.9 & 56.5 & 26.1 & 40.6  & \cellcolor[HTML]{FFFFED}3.6 & 38.7 & 23.7 & 33.7 & 33.6 & 26.3 & 31.8 \\

   OW-DETR~\cite{Owdetr} & \cellcolor[HTML]{FFFFED}5.7  & 71.5 & \cellcolor[HTML]{FFFFED}6.2 & 62.8 & 27.5 & 43.8 & \cellcolor[HTML]{FFFFED}6.9 & 45.2 & 24.9 & 38.5 & 38.2 & 28.1 & 33.1 \\
   
   CAT$^ \dag$~\cite{CAT} & \cellcolor[HTML]{FFFFED} 24.0  & \textbf{74.2} & \cellcolor[HTML]{FFFFED} 23.0 & \textbf{67.6} & 35.5 & 50.7 & \cellcolor[HTML]{FFFFED} 24.6 & \textbf{51.2} & 32.6 & \textbf{45.0} & \textbf{45.4} & \textbf{35.1} & \textbf{42.5} \\

   PROB~\cite{prob} & \cellcolor[HTML]{FFFFED} 17.6  & 73.4 & \cellcolor[HTML]{FFFFED} 22.3 & 66.3 & 36.0 & 50.4 & \cellcolor[HTML]{FFFFED} 24.8 & 47.8 & 30.4 & 42.0 & 42.6 & 31.7 & 39.9 \\

   \textbf{Ours: USD} \texttt{-ASF} & \cellcolor[HTML]{FFFFED} 19.2  & 72.9 & \cellcolor[HTML]{FFFFED} 22.4 & 64.9 & \textbf{38.9} & \textbf{51.2} & \cellcolor[HTML]{FFFFED} 25.4 & 50.1 & \textbf{34.7} & \textbf{45.0} & 43.4 & 33.6 & 40.9 \\

   \textbf{Ours: USD $^ \ddag$} & \cellcolor[HTML]{FFFFED} \textbf{51.1}  & 69.8 & \cellcolor[HTML]{FFFFED} \textbf{52.1} & 60.9 & 33.8 & 46.7 & \cellcolor[HTML]{FFFFED} \textbf{49.9} & 45.8 & 30.2 & 40.6 & 39.0 & 28.7 & 36.4 \\
   \bottomrule
   
   \end{tabular}%
   }
   \label{table:t1_owod}
   \vspace{-3mm}
\end{table*}

\begin{table*}[t]

   \centering
   \caption{\textbf{The impact of our contributions on the baseline model}. 
   USD \texttt{-ASF} and USD \texttt{-DOL} are our model without auxiliary supervision framework (ASF) and decoupled objectness learning (DOL).  
   To provide a comprehensive analysis, we include the performance of DDETR and an upper bound DDETR, which is trained using ground-truth unknown class annotations, as reported in~\cite{Owdetr}.
   }
   \label{tab:ablation_table}
   \setlength{\tabcolsep}{2pt}
   \adjustbox{width=\textwidth}{
   \begin{tabular}{@{}l|cc|cccc|cccc|ccc@{}}
   \toprule

    \textbf{Task IDs} ($\rightarrow$)& \multicolumn{2}{c|}{\textbf{Task 1}} & \multicolumn{4}{c|}{\textbf{Task 2}} & \multicolumn{4}{c|}{\textbf{Task 3}} & \multicolumn{3}{c}{\textbf{Task 4}} \\ \midrule
    
    & \cellcolor[HTML]{FFFFED}{U-Recall} & \multicolumn{1}{c|}{\cellcolor[HTML]{EDF6FF}{mAP ($\uparrow$)}} & \cellcolor[HTML]{FFFFED}{U-Recall} & \multicolumn{3}{c|}{\cellcolor[HTML]{EDF6FF}{mAP ($\uparrow$)}} & \cellcolor[HTML]{FFFFED}{U-Recall} & \multicolumn{3}{c|}{\cellcolor[HTML]{EDF6FF}{mAP ($\uparrow$)}} & \multicolumn{3}{c}{\cellcolor[HTML]{EDF6FF}{mAP ($\uparrow$)}}  \\

    & \cellcolor[HTML]{FFFFED}($\uparrow$) & \begin{tabular}[c]{@{}c}Current \\ known\end{tabular} & \cellcolor[HTML]{FFFFED}($\uparrow$) & \begin{tabular}[c]{@{}c@{}}Previously\\  known\end{tabular} & \begin{tabular}[c]{@{}c@{}}Current \\ known\end{tabular} & Both & \cellcolor[HTML]{FFFFED}($\uparrow$) & \begin{tabular}[c]{@{}c@{}}Previously \\ known\end{tabular} & \begin{tabular}[c]{@{}c@{}}Current \\ known\end{tabular} & Both & \begin{tabular}[c]{@{}c@{}}Previously \\ known\end{tabular} & \begin{tabular}[c]{@{}c@{}}Current \\ known\end{tabular} & Both \\ \midrule

   Upper Bound & \cellcolor[HTML]{FFFFED}31.6  & 62.5  & \cellcolor[HTML]{FFFFED}40.5 & 55.8 & 38.1 & 46.9 & \cellcolor[HTML]{FFFFED}42.6 & 42.4 & 29.3 & 33.9 & 35.6 & 23.1 & 32.5 \\

   \begin{tabular}[c]{@{}l@{}} DDETR \cite{ddetr} \end{tabular} & -&60.3  & - & 54.5 & 34.4 & 44.7  & - & 40.0 & 17.7 & 33.3 & 32.5 & 20.0 & 29.4 \\ 
   \midrule
   \midrule

   \textbf{USD}\texttt{-ASF}   & \cellcolor[HTML]{FFFFED} 21.6  & \textbf{59.9 }& \cellcolor[HTML]{FFFFED} 19.7 & \textbf{56.6} & \textbf{32.5} & \textbf{44.6}  & \cellcolor[HTML]{FFFFED} 23.5 & \textbf{43.5} & \textbf{21.9} & \textbf{36.3} & \textbf{35.4} & \textbf{18.9} & \textbf{31.3} \\

   \textbf{USD}\texttt{-DOL}  & \cellcolor[HTML]{FFFFED} 33.1  & 57.0 & \cellcolor[HTML]{FFFFED} 32.7 & 54.5 & 30.9 & 42.7  & \cellcolor[HTML]{FFFFED} 32.7 & 42.2 & 19.7 & 34.7 & 34.7 & 16.6 & 30.1 \\

   Final: \textbf{USD} & \cellcolor[HTML]{FFFFED} \textbf{36.1}  & 58.4& \cellcolor[HTML]{FFFFED} \textbf{34.7} & 54.3 & 31.4 & 42.7  & \cellcolor[HTML]{FFFFED} \textbf{33.3} & 41.5 & 20.5 & 34.5 & 33.4 & 16.4 & 29.2 \\ 
   \bottomrule
   
   \end{tabular}%
   }
   
   \end{table*}

\begin{figure*}
   \centerline{\includegraphics[width=1.0\linewidth]{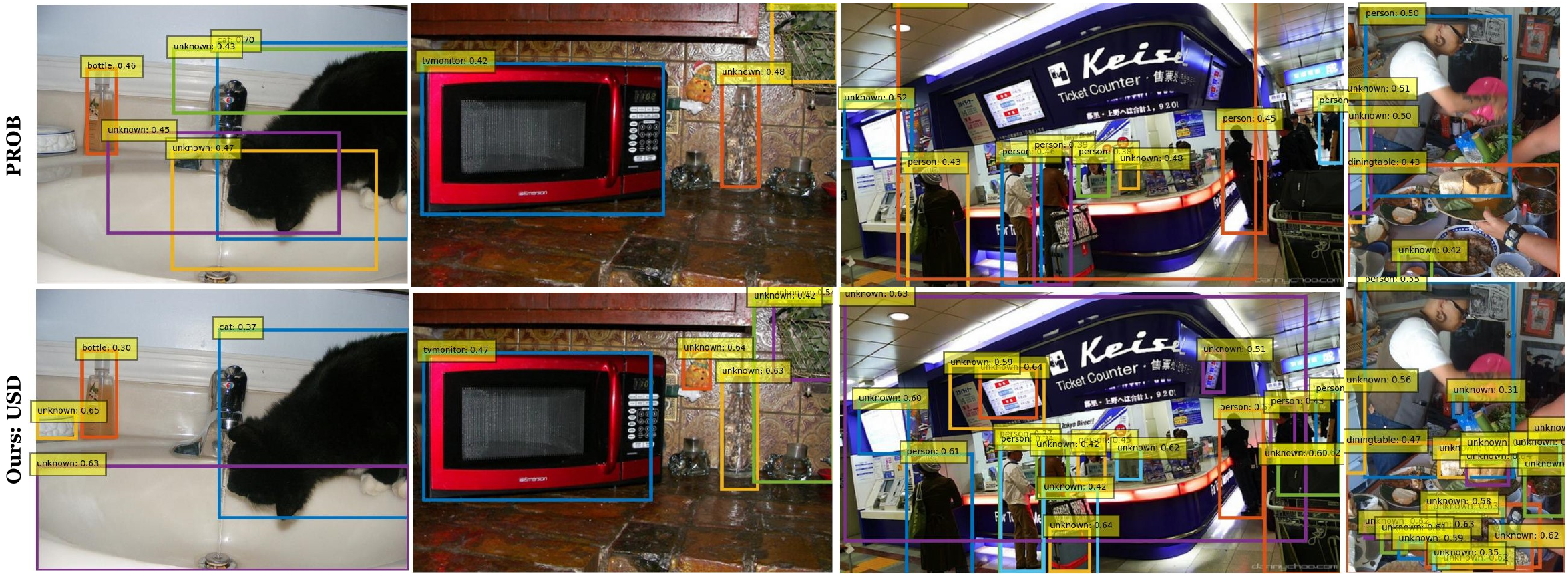}}
   \caption{Qualitative results on example images from MS-COCO test set.
   }
   \label{fig:qualitative_results}
   \vspace{-3mm}
\end{figure*}

\subsection{Comparison With State-of-the-art Methods}

We conducted a comprehensive comparison of our proposed USD with other state-of-the-art OWOD methods on the M-OWODB and S-OWODB benchmarks, as presented in Table~\ref{table:t1_owod}.
ORE~\cite{towards} relies on a held-out test set of unknown images to estimate unknown energy, leakaging the information of test set, as illustrated in ~\cite{Owdetr,revisiting}.
Therefore, we add a comparison variant called ORE\texttt{-EBUI}, which removes the energy based unknown identifier (EBUI).
To make the comparison fair, we additionally category existing methods into which add additional calculations marked by `` $\dag$'' and which add additional data or models marked by `` $\ddag$''.
CAT, for instance, incorporates a cascade decoder that requires running twice during inference, resulting in a noticeable increase in computation.
To account for this, we include a comparison variant called CAT\texttt{-cddw}, which removes the cascade decoder and exhibits a similar level of computation as OW-DETR~\cite{Owdetr}, PROB~\cite{prob}, and our USD.
Furthermore, our USD leverages additional knowledge from SAM, so we provide a comparison variant called USD\texttt{-ASF}, which removes ASF part of USD.

\textbf{M-OWODB Comparison:} As presented in the first section of Tab.~\ref{table:t1_owod}, our USD achieves a remarkable improvement in unknown object detection, surpassing the baseline PROB by more than 15 points in U-Recall.
Additionally, there is only a minor decline of 1.1-2.3 points in mAP across all four tasks, which is reasonable because more unknown objects are detected, increasing the difficulty of correctly classifying known classes.
Furthermore, USD has a stronger capability to prevent the detection of unknown classes as known classes, as shown in Tab.~\ref{tab:wi_ose}.
For instance, USD achieves A-OSE values of 3137, 1677, and 933, which are superior to PROB's values of 5195, 6452, and 2641, in Task 1-3.
Compared to OW-DETR~\cite{Owdetr}, the pioneering transformer-based detector, USD achieves a substantial improvement of 4.8$\times$, 5.6$\times$, and 5.8$\times$ in U-Recall for Tasks 1-3, respectively.
Moreover, when compared to the recent state-of-the-art (SOTA) method CAT$^\dag$, which involves additional calculations during inference, USD outperforms it by 14.3, 15.5, and 8.9 points in U-Recall for Tasks 1-3, respectively.
To ensure a fair comparison, we evaluate USD\texttt{-ASF} against all other methods that do not introduce additional computation or additional data/model-assisted supervision, including OW-DETR, 2B-OCD, CAT\texttt{-cddw}, and PROB. 
Notably, USD\texttt{-ASF} achieves the highest U-Recall and surpasses PROB by 2.2, 2.3, and 3.9 points for Tasks 1-3, while maintaining a close mAP in known object detection.
The significant improvement in U-Recall achieved by USD\texttt{-ASF} highlights the criticality of addressing the conflict in learning objectness and classification boundaries.

\textbf{S-OWODB Comparison:} 
As indicated in the second section of Table~\ref{table:t1_owod}, our USD achieves a notable improvement in U-Recall, surpassing the recent SOTA method CAT$\dag$ by 27.1, 29.1, and 25.3 points in Task 1-3, respectively.
This result clearly highlights the significant benefits of incorporating SAM and our proposed ASF, which greatly enhance the detection of unknown objects.
However, there is a slight decrease in mAP (approximately 3 points) across Task 1-4 when compared to PROB.
This trade-off in performance is a consequence of prioritizing the improvement in U-Recall. 
The additional unknown supervision makes it more challenging for the model to discriminate known objects.
Furthermore, when compared to other methods without $\dag$ and $\ddag$ symbols, USD\texttt{-ASF} achieves the best U-Recall and mAP values.
Specifically, USD\texttt{-ASF} outperforms PROB in terms of U-Recall by 1.6, 0.1, and 0.6 points in Task 1-3, and in terms of mAP by 0.8, 3.0, and 1.0 points in Task 2-4.
In S-OWODB, our proposed DOL not only exhibits a improvement in U-Recall, similar to the results obtained in M-OWODB, but also achieves an improvement in mAP.

\textbf{Qualitative Results:}
Fig.~\ref{fig:qualitative_results} depicts the qualitative results of both PROB and our proposed USD.
This visualization vividly demonstrate the enhanced capability of USD in detecting unknown objects, which aligns with the quantitative results presented in Tab.~\ref{table:t1_owod}. 
Whether the scene is relatively simple, as portrayed in the first two images, or more complex, as depicted in the last two images, USD showcases its comprehensive ability to detect unknown objects.
This consistent performance across various scenes further highlights the effectiveness of our method.

\subsection{Ablation Study}
All ablation studies are conducted on M-OWODB. 
The experiments of Tab.~\ref{tab:decoder}, Tab.~\ref{tab:asf}, and Tab.~\ref{tab:pseudo} are conducted in Task 1 of M-OWODB.

\begin{table}
    \centering
    \caption{Effect on learning objectness and classification boundaries in different decoder layers. We remove auxiliary losses in DDETR to isolate the effects of them. }
    \label{tab:decoder}
    {%W
    \scalebox{0.9}{
    \begin{tabular}{@{}ccc@{}}
    \toprule
    { Decoder Layer} & U-Recall & mAP \\ \midrule
 
    1  & \textbf{22.69}  & 55.65  \\
    
    2 & 22.54  & 55.20 \\ 
    
    3 & 22.34 & 55.05  \\ 
 
    4 & 21.39 & \textbf{55.80} \\ 
 
    5 & 20.27 & 55.38  \\ 
 
    6 & 19.42 & 54.50 \\ 
    
    \bottomrule
 \end{tabular}}
 }
 \vspace{-2mm}
 \end{table}

 \begin{table}
    \centering
    \caption{Effect of pseudo-labeling (PL) and soft-weighting (SW) strategies in ASF. }
    \label{tab:asf}
    {%
    \scalebox{0.9}{
    \begin{tabular}{@{}ccccc@{}}
    \toprule
    SAM & PL & SW & U-Recall & mAP \\ \midrule
 
    \checkmark &  &   & 30.98  & 55.02  \\
    
    \checkmark & \checkmark &  & 32.49  & 54.62 \\ 
 
    \checkmark & \checkmark & \checkmark & \textbf{36.08} & \textbf{58.39} \\

    \bottomrule
 \end{tabular}}
 }
 \vspace{-2mm}
 \end{table}
 
 \begin{table}
    \centering
    \caption{Effect of different pseudo-labeling thresholds in ASF. }
    \label{tab:pseudo}
    {%
    \scalebox{0.9}{
    \begin{tabular}{@{}ccc@{}}
    \toprule
    { Threshold} & U-Recall & mAP \\ \midrule
 
    0.6  & 35.76  & 56.72  \\
    
    0.7 & \textbf{36.08}  & \textbf{58.39} \\ 
    
    0.8 & 34.23  & 57.01 \\ 
    \bottomrule
 \end{tabular}}
 }
 \vspace{-2mm}
 \end{table}

\textbf{Component Ablation Study:}
Tab.~\ref{tab:ablation_table} shows the contributions of our proposed ASF and DOL.
It is evident that ASF plays a crucial role in improving the detection of unknown objects, significantly enhancing the U-Recall. 
Additionally, DOL effectively mitigates the conflict problem in learning objectness and classification boundaries, resulting in improved detection of both unknown and known objects.
When comparing USD with the upper bound DDETR, which uses the unknown groundtruth to guide the learning of unknown object detection, USD achieves a close U-Recall value to it. 
This demonstrates the feasibility of utilizing SAM to address the challenge of limited unknown object annotation in OWOD, providing a promising solution to the data annotation dilemma problem.

\textbf{Separated Learning in DOL:} We performed a comprehensive study to evaluate the impact of separately learning the objectness and classification boundaries.
We assigned the learning of objectness to different decoder layers while maintaining the classification learning in the last decoder layer.
The results, as depicted in Tab.~\ref{tab:decoder}, clearly indicate that learning objectness in Layer 1 achieves the highest U-Recall with a satisfactory mAP. 
Conversely, Layer 6 exhibits the poorest performance in terms of U-Recall and mAP. 
These findings emphasize the potential conflict that arises when simultaneously learning objectness and classification boundaries, leading to suboptimal performance.

\textbf{Pseudo-labeling and soft-weighting strategies in ASF:} 
As shown in Tab.~\ref{tab:asf}, we compared the results obtained by directly using the SAM's output with those achieved by exploiting pseudo-labeling and soft-weighting strategies in ASF.
The inclusion of pseudo-labeling and soft-weighting strategies led to significant improvements in both U-Recall and mAP, highlighting the importance of addressing the noise present in SAM's output.
Additionally, we conducted a sensitivity analysis of the pseudo-labeling threshold, as shown in Tab.~\ref{tab:pseudo}, and found that a threshold of 0.7 yielded the best results.

\section{Conclusions}
This paper focus on enhancing the detection of unknown objects in Open-World Object Detection. 
We begin by identifying a crucial conflict issue in learning objectness and classification boundaries, and propose a solution that outperforms the recent state-of-the-art method, PROB, in terms of both unknown and known object detection.
Then, to tackle the annotation dilemma in unknown objects, we leverage the power of a Large Vision Model (LVM), the Segment Anything Model (SAM), to identify potential unknown objects in open-world. 
Additionally, we introduce a generic auxiliary supervision framework that combines pseudo-labeling and soft-labeling techniques to mitigate the impact of noise in the output of SAM.
Our proposed method, USD, significantly advances the field of unknown object detection, promoting the development of this field.  

{\small
\bibliographystyle{ieee_fullname}
\bibliography{egbib}
}

\clearpage
\appendix
% \onecolumn
\begin{minipage}[t]{\columnwidth}
\centering
\Large
\bf
Supplementary Materials
\end{minipage}

\section{Detailed Illustration of Data Splits}
\label{sec:sup:data}
Tab.~\ref{table6} provides detailed information regarding the class distribution, number of images, and instances in both the M-OWODB and S-OWODB data splits.
In M-OWODB, the dataset is divided into four tasks, each focusing on a specific set of classes. 
Task 1 consists of the 20 Pascal VOC classes, while Tasks 2 to 4 encompass the remaining 60 classes from MS COCO. 
This division of classes introduces a potential concern of information leakage, as different classes belonging to the same superclass are distributed across different tasks.
Consequently, it becomes relatively easier to detect unknown objects within the same superclass.
To mitigate this issue, the S-OWODB data split adopts a more explicit separation of superclasses. 
For instance, superclasses such as animals and vehicles, sports and food are kept distinct from one another. 
Similar to M-OWODB, S-OWODB also consists of four tasks. 
However, in S-OWODB, all the classes belonging to a particular superclass are introduced together within a single task.
This approach ensures a clearer demarcation between different superclasses in the dataset.

\begin{table*}[h]
    \centering
    \renewcommand\arraystretch{1.25}
    \caption{\textbf{The details of M-OWODB and S-OWODB data splits.} } 
    \resizebox{0.7\linewidth}{!}{
    \begin{tabular}{lcccc}
    \toprule
     \textbf{Task IDs}& \textbf{Task 1}   & \textbf{Task 2 } & \textbf{Task 3}     & \textbf{Task 4}   \\ \midrule \midrule
    \multicolumn{5}{c}{ \textbf{M-OWODB Data Split}}\\\midrule
    \multicolumn{1}{l|}{ \textbf{Semantic Split}}      & \multicolumn{1}{c}{\begin{tabular}[c]{@{}c@{}}VOC \\ Classes\end{tabular}} & \multicolumn{1}{c}{\begin{tabular}[c]{@{}c@{}}Outdoor, Accessories, \\ Appliances, Truck\end{tabular}} & \multicolumn{1}{c}{\begin{tabular}[c]{@{}c@{}}Sports, \\ Food\end{tabular}} & \begin{tabular}[c]{@{}c@{}}Electronic, Indoor, \\ Kitchen, Furniture\end{tabular} \\ \midrule
    \multicolumn{1}{l|}{\# train images} & \multicolumn{1}{c}{16551}    & \multicolumn{1}{c}{45520}   & \multicolumn{1}{c}{39402}  & 40260    \\
    \multicolumn{1}{l|}{\# test images}     & \multicolumn{1}{c}{4952}    & \multicolumn{1}{c}{1914}   & \multicolumn{1}{c}{1642}    & 1738   \\
    \multicolumn{1}{l|}{\# train instances} & \multicolumn{1}{c}{47223}   & \multicolumn{1}{c}{113741}   & \multicolumn{1}{c}{114452}   & 138996      \\
    \multicolumn{1}{l|}{\# test instances}  & \multicolumn{1}{c}{14976}   & \multicolumn{1}{c}{4966}   & \multicolumn{1}{c}{4826}  & 6039  \\  \midrule\midrule
    \multicolumn{5}{c}{ \textbf{S-OWODB Data Split}} \\\midrule
    \multicolumn{1}{l|}{\textbf{Semantic Split}}      & \multicolumn{1}{c}{\begin{tabular}[c]{@{}c@{}}Animals,Person, \\ Vehicles\end{tabular}} & \multicolumn{1}{c}{\begin{tabular}[c]{@{}c@{}}Appliances, Accessories, \\ Outdoor, Furniture\end{tabular}} & \multicolumn{1}{c}{\begin{tabular}[c]{@{}c@{}}Sports, \\ Food\end{tabular}} & \begin{tabular}[c]{@{}c@{}}Electronic, Indoor, \\ Kitchen\end{tabular} \\ \midrule
    \multicolumn{1}{l|}{\# train images} & \multicolumn{1}{c}{89490}    & \multicolumn{1}{c}{55870}   & \multicolumn{1}{c}{39402}  & 38903    \\
    \multicolumn{1}{l|}{\# test images}     & \multicolumn{1}{c}{3793}    & \multicolumn{1}{c}{2351}   & \multicolumn{1}{c}{1642}    & 1691   \\
    \multicolumn{1}{l|}{\# train instances} & \multicolumn{1}{c}{421243}   & \multicolumn{1}{c}{163512}   & \multicolumn{1}{c}{114452}   & 160794      \\
    \multicolumn{1}{l|}{\# test instances}  & \multicolumn{1}{c}{17786}   & \multicolumn{1}{c}{7159}   & \multicolumn{1}{c}{4826}  & 7010   \\ \bottomrule 
    \end{tabular}}
    \label{table6}
\end{table*}

\begin{table*}[t]
    \centering
    \caption{ \textbf{Open-set Performance Comparison on M-OWODB.} Wilderness impact (WI), absolute open set error (A-OSE) and unknown class recall (U-Recall) are the metrics to meansure to open-set performance.
    WI and A-OSE indicates the relative and absolute error in classifying unknown objects into known classes, and they are the smaller the better. 
    U-Recall represents the capability of unknown object detection, and it is the bigger the better.}
    \label{tab:wi_ose}
    \setlength{\tabcolsep}{10pt}
    \adjustbox{width=\textwidth}{
    \begin{tabular}{@{}l|ccc|ccc|ccc}
    \toprule
     \textbf{Task IDs} ($\rightarrow$)& \multicolumn{3}{c|}{\textbf{Task 1}} & \multicolumn{3}{c|}{\textbf{Task 2}} & \multicolumn{3}{c}{\textbf{Task 3}} \\
    \midrule

     & \cellcolor[HTML]{FFFFED}{U-Recall} & \cellcolor[HTML]{EDF6FF}{WI} & \cellcolor[HTML]{EDF6FF}{A-OSE} & \cellcolor[HTML]{FFFFED}{U-Recall} & \cellcolor[HTML]{EDF6FF}{WI} & \cellcolor[HTML]{EDF6FF}{A-OSE}  & \cellcolor[HTML]{FFFFED}{U-Recall} & \cellcolor[HTML]{EDF6FF}{WI} & \cellcolor[HTML]{EDF6FF}{A-OSE} \\

     & \cellcolor[HTML]{FFFFED}($\uparrow$) & \cellcolor[HTML]{EDF6FF}($\downarrow$) & \cellcolor[HTML]{EDF6FF}($\downarrow$) & \cellcolor[HTML]{FFFFED}($\uparrow$) & \cellcolor[HTML]{EDF6FF}($\downarrow$) & \cellcolor[HTML]{EDF6FF}($\downarrow$) & \cellcolor[HTML]{FFFFED}($\uparrow$) & \cellcolor[HTML]{EDF6FF}($\downarrow$) & \cellcolor[HTML]{EDF6FF}($\downarrow$) \\
    
     \midrule
    
    % \midrule
    
    % edf6ff

    ORE\texttt{-EBUI}~\cite{towards} & \cellcolor[HTML]{FFFFED}4.9  & 0.0621 & 10459 & \cellcolor[HTML]{FFFFED}2.9 & 0.0282 & 10445 & \cellcolor[HTML]{FFFFED}3.9 & 0.0211 & 7990  \\

    2B-OCD ~\cite{wu2022two} & \cellcolor[HTML]{FFFFED}12.1 & 0.0481 & - & \cellcolor[HTML]{FFFFED}9.4 & 0.160 & - & \cellcolor[HTML]{FFFFED}11.6 & 0.0137 & -  \\

    OW-DETR \cite{Owdetr} & \cellcolor[HTML]{FFFFED}7.5  & 0.0571 & 10240 & \cellcolor[HTML]{FFFFED}6.2 & 0.0278 & 8441 & \cellcolor[HTML]{FFFFED}5.7 & 0.0156 & 6803  \\
    
    OCPL ~\cite{OCPL} & \cellcolor[HTML]{FFFFED} 8.3  & \textbf{0.0413} & 5670 & \cellcolor[HTML]{FFFFED} 7.6 & \textbf{0.0220} & 5690 & \cellcolor[HTML]{FFFFED}11.9 & 0.0162 & 5166  \\
 
    PROB~\cite{prob} & \cellcolor[HTML]{FFFFED} 19.4  & 0.0569 & 5195 & \cellcolor[HTML]{FFFFED} 17.4 & 0.0344 & 6452 & \cellcolor[HTML]{FFFFED} 19.6 & 0.0151 &2641  \\
    \midrule
    \midrule
    
    \textbf{Ours: USD} & \cellcolor[HTML]{FFFFED} \textbf{36.1}  & 0.0544 & \textbf{3137} & \cellcolor[HTML]{FFFFED} \textbf{34.7} & 0.0241 & \textbf{1677}& \cellcolor[HTML]{FFFFED} \textbf{33.3} & \textbf{0.0130} & \textbf{933}  \\
    \bottomrule
    
    \end{tabular}%
    }
    
 \end{table*}

\section{Additional Experimentaion Details}
\label{sec:sup:detail}
Our experimental setup closely follows that of PROB~\cite{prob} for consistency and comparison purposes. 
The training of our USD model is performed on four Nvidia A100 40GB GPUs, with a batch size of 5 per GPU. 
We utilize the Adam optimizer with $\beta_1=0.9$, $\beta_2=0.999$, and a weight decay of $10^{-4}$.
The initial learning rate is set to $2 \times 10^{-3}$ and reduced by 10 after 35 epochs.
During the incremental learning step, the learning rate is set to $2 \times 10^{-4}$.
We maintain a set of 50 stored exemplars per known class for this process.

Regarding the Segment Anything Model (SAM), we employ the \texttt{sam\_vit\_h} model for inference. 
In the ``everything'' mode, we configure a $32 \times 32$ point grid. 
To assess the quality of predicted masks, we set the prediction IoU threshold to 0.95. 
This threshold helps estimate the overlap between the predicted masks and the corresponding ground truth masks, ensuring high accuracy in the segmentation results.
We also set the stability score threshold to 0.95 to ensure reliable predictions.
Lastly, the minimum mask region area is set to 200 to filter out small regions.

Additionally, in our post-processing procedure, we apply a geometric mean of objectness and classification predictions to obtain more accurate results, formalized as:
\begin{equation}
    P_s = P_{obj} ^ \gamma \cdot (P_{cls}) ^ {1- \gamma}.
    \label{eq:gmean}
\end{equation}
We set $\gamma$ to 0.6 in M-OWODB and 0.7 in S-OWODB.

%\clearpage
\begin{figure}
    \centerline{\includegraphics[width=1.0\linewidth]{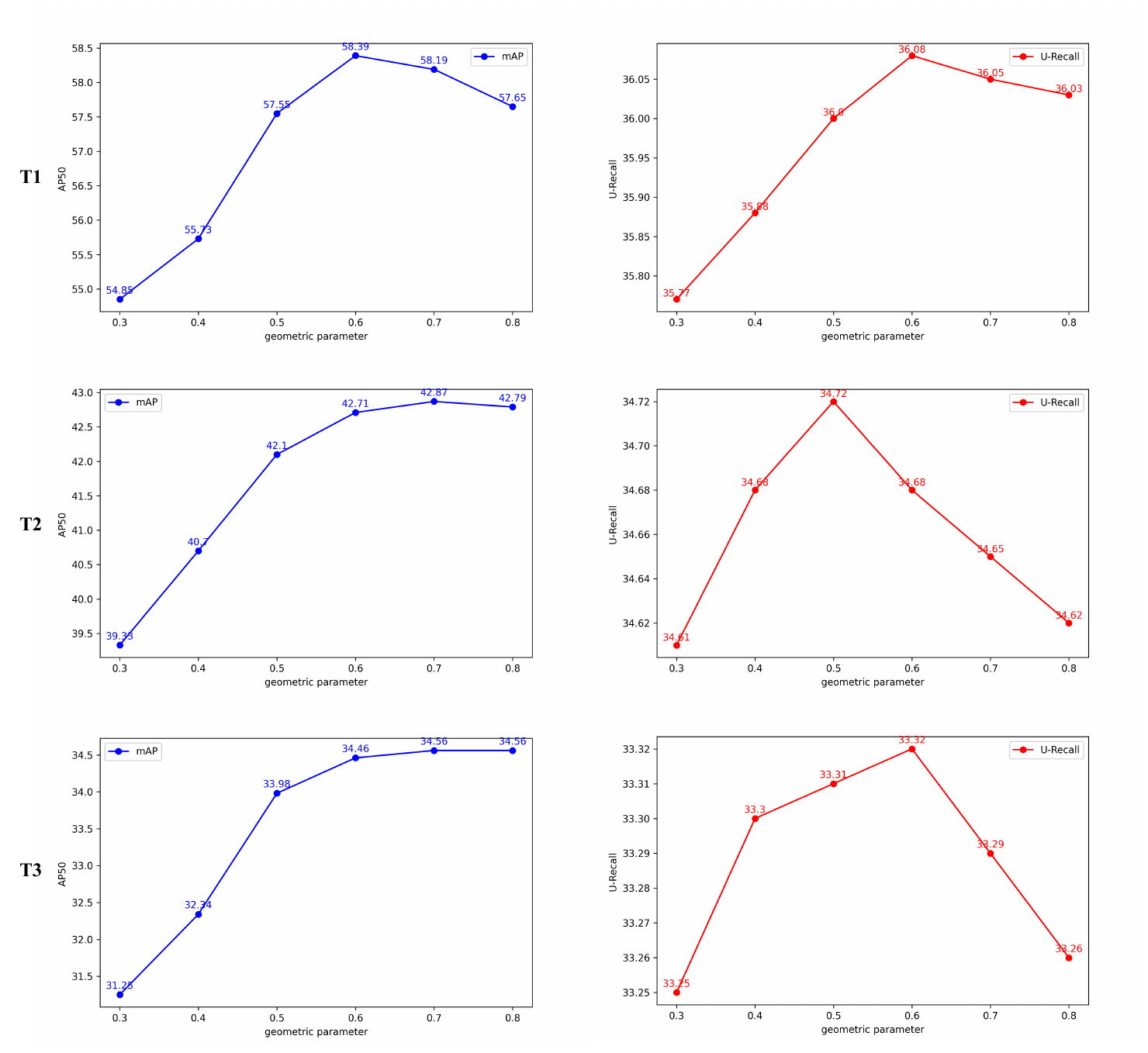}}
    \caption{\textbf{Geometric Parameter Analysis on M-OWODB.} We set different $\lambda$ in Eq.~\ref{eq:gmean} in inference and analyze the U-Recall and mAP values.
    U-Recall and mAP achieve the peak at $\lambda$ of 0.6-0.7. 
    }
    \label{fig:m-zhexian}
    \vspace{-3mm}
 \end{figure}

 \begin{figure}
    \centerline{\includegraphics[width=1.0\linewidth]{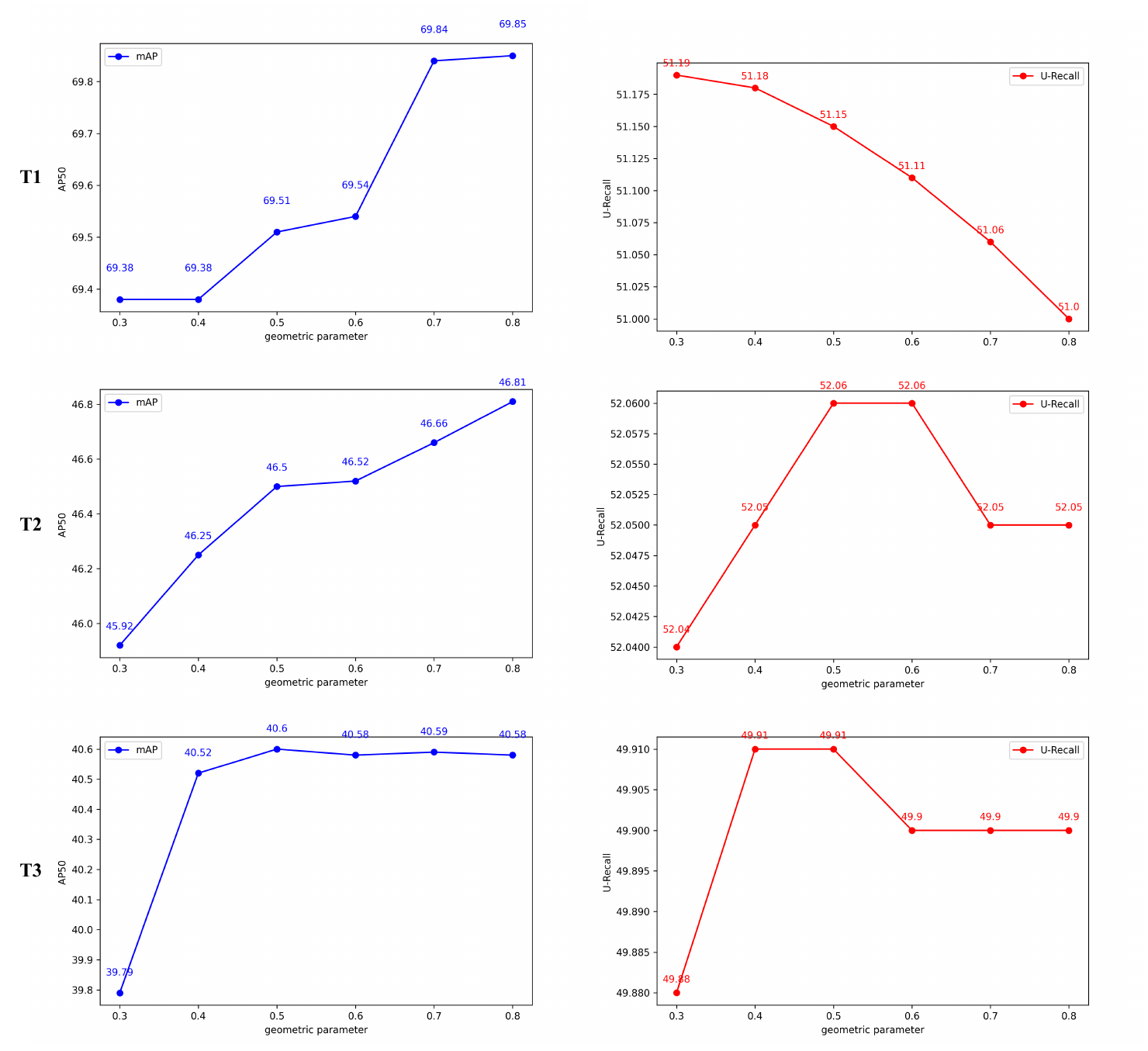}}
    \caption{\textbf{Geometric Parameter Analysis on S-OWODB.} We set different $\lambda$ in Eq.~\ref{eq:gmean} in inference and analyze the U-Recall and mAP values.
    mAP achieves the peak at $\lambda$ of 0.7-0.8 and U-Recall achieves the peak at $\lambda$ of 0.6-0.7. 
    }
    \label{fig:s-zhexian}
    \vspace{-3mm}
 \end{figure}

\section{Open-set Performance Comparison}
\label{sec:sup:openset}
In the context of Open-World Object Detection (OWOD), the performance of detecting unknown objects is evaluated using the WI and A-OSE metrics, as mentioned in~\cite{towards}.

WI (Wilderness Impact) is a metric that measures the proportion of unknown objects incorrectly classified as known classes. 
WI is formulated as follows:
\begin{equation}
    \mathrm{WI}=\frac{P_{K}}{P_{{K} \cup {U}}}-1
\end{equation}where $P_{K}$ is precision of known classes and $P_{{K} \cup {U}}$ is the precision of both known classes and unknown classes (remaining classes in MS COCO).
WI reflects the relative error in classifying unknown objects as known classes.

A-OSE (Absolute Open-Set Error) measures the total number of unknown objects detected as known classes. 
It represents the absolute error in detecting and classifying unknown objects.

In Tab.~\ref{tab:wi_ose}, it can be observed that USD achieves a significant improvement in U-Recall, with an increase of over 15 points in Task 1-3. 
Furthermore, the open-set errors, especially the A-OSE, are reduced compared to the PROB method~\cite{prob}.
A-OSE is found to be 0.6 $\times$, 2.8 $\times$, and 1.8 $\times$ lower in USD compared to PROB in Task 1-3. 
This result demonstrates that USD not only detects more unknown objects but also possesses enhanced capabilities in preventing the misclassification of unknown objects as known classes.

\section{Additional Ablation Study}
\textbf{Different $\lambda$ in Eq.~\ref{eq:gmean}:}
As shown in Fig.~\ref{fig:m-zhexian} and Fig.~\ref{fig:s-zhexian}, we analyze the results of U-Recall and mAP in different values of $\lambda \in [0, 1]$ in Task 1-3 on M-OWODB and S-OWODB data splits.
The changes in $\lambda$ have a greater effect on mAP and a smaller effect on U-Recall.
On M-OWODB, it is evident that U-Recall and mAP peak at the $\lambda$ of 0.6-0.7.
Therefore, we pick $\lambda=0.6$ on M-OWODB.
On S-OWODB, the pick of $\lambda$ is not clear, mAP achieves the peak at 0.7-0.8 while U-Recall achieves the peak at 0.6-0.7.
Therefore, we pick $\lambda=0.7$ on M-OWODB.
When we set $\lambda=0$, i.e., only using $P_{cls}$ to represent $P_s$, the mAP is droped to 53.86 in Task 1 on M-OWODB, which indicates that the addition of $P_{obj}$ helps the improvement of mAP.

\begin{table}
    \centering
    \caption{Experiments on $\alpha$ in Eq.~\ref{eq:score}}
    \label{tab:pseudo-alpha}
    {%
    \scalebox{0.9}{
    \begin{tabular}{@{}ccc@{}}
    \toprule
     $\alpha$ & U-Recall & mAP \\ \midrule
 
    0.4  & 35.44  & 57  \\
    
    0.5 & \textbf{36.08}  & \textbf{57.55} \\ 
    
    0.6 & 35.99  & 56.53 \\ 
    \bottomrule
 \end{tabular}}
 }
\end{table}

\begin{table}
    \centering
    \caption{Experiments on loss coefficients}
    \label{tab:loss-coe}
    {%
    \scalebox{0.9}{
    \begin{tabular}{@{}cccc@{}}
    \toprule
    $\lambda_{obj}^{pse}$ & $\lambda_{reg}^{pse}$ & U-Recall & mAP \\ \midrule
    0 & 5  & 29.79 & 56.62  \\
    4e-4 & 5  & 35.57 & 55.89  \\
    8e-4 & 5  & 35.93 & 56.86  \\
    8e-5 & 5  & \textbf{36.08} & 57.55  \\
    8e-5 & 2.5  & 35.19 & 57.32  \\
    8e-5 & 0.5  & 31.85 & \textbf{58.45}  \\

    \bottomrule
 \end{tabular}}
 }
\end{table}

\textbf{Different $\alpha$ in Eq.~\ref{eq:score}:}
Table~\ref{tab:pseudo-alpha} presents the results obtained by using different values of $\alpha$ in Eq.~\ref{eq:score} to calculate the scores.
Notably, among the range of values tested, $\alpha=0.5$ emerges as the optimal choice.
This value serves as a parameter to strike a balance between the predictions generated by the model and the auxiliary model.
The selection of $\alpha$ hinges upon the detection capabilities of the auxiliary model and the learning situations of the primary model.

\textbf{Loss coefficient experiments:}
In Tab.~\ref{tab:loss-coe}, we present a comparison of U-Recall and mAP using different loss coefficients.
By varying the $\lambda_{obj}^{pse}$ and $\lambda_{obj}^{reg}$, we investigate their impact on performance.
When $\lambda_{obj}^{pse}$ is set to $8 \times 10^{-5}$, which is one-tenth of the coefficient used in $\mathcal{L}_{obj}$, we achieve the best U-Recall and mAP values. 
This indicates that a slightly reduced weight on the objectness loss in the auxiliary supervision yields optimal results.
On the other hand, when $\lambda_{obj}^{reg}$ is set to 0.5, which is one-tenth of the coefficient used in $\mathcal{L}_{reg}$, we obtain the best mAP but an unsatisfactory U-Recall. 
This suggests that reducing the weight on the regression loss in the auxiliary supervision prioritizes accurate localization but may compromise the detection of unknown objects.
Interestingly, when $\lambda_{obj}^{reg}$ is increased to 5, we observe a notable improvement of 4.28 points in U-Recall with a modest decrease of only 0.9 points in mAP. 
This trade-off between U-Recall and mAP leads us to select $\lambda_{obj}^{reg}=5$ as a reasonable choice.
Based on the results presented in Tab.~\ref{tab:loss-coe}, it is evident that both objectness and regression learning in the auxiliary supervision contribute to performance improvement. 
The U-Recall is particularly influenced by the change in $\lambda_{obj}^{reg}$, whereas the change in $\lambda_{obj}^{pse}$ tends to affect the mAP.
This phenomenon suggests that $\mathcal{L}_{reg}^{pse}$ encourages the model to detect more potential unknown objects, while $\mathcal{L}_{obj}^{pse}$ enhances the model's objectness estimation, thereby promoting accurate classification.

\section{Additional Qualitative Results}
\textbf{Visualization Setting Details:}
Due to the inherent challenge of defining and annotating all unknown objects, the evaluation of unknown object detection typically relies on the metric of Recall rather than mAP.
This metric choice incentivizes existing methods to prioritize recall improvement without considering the accuracy of unknown object detection. 
As a result, these methods tend to generate numerous dense unknown boxes throughout the image, leading to poor visualization quality.
To facilitate a fair visualization comparison, we adopt the approach used in PROB~\cite{prob}.
We define the top-100 predictions and visualize the unknown predictions overlapped with the ground truth boxes.
In this way, the visualization setting is transformed to detect more unknown ground-truths with the limited number of predictions.

\begin{figure*}
    \centerline{\includegraphics[width=0.86\linewidth]{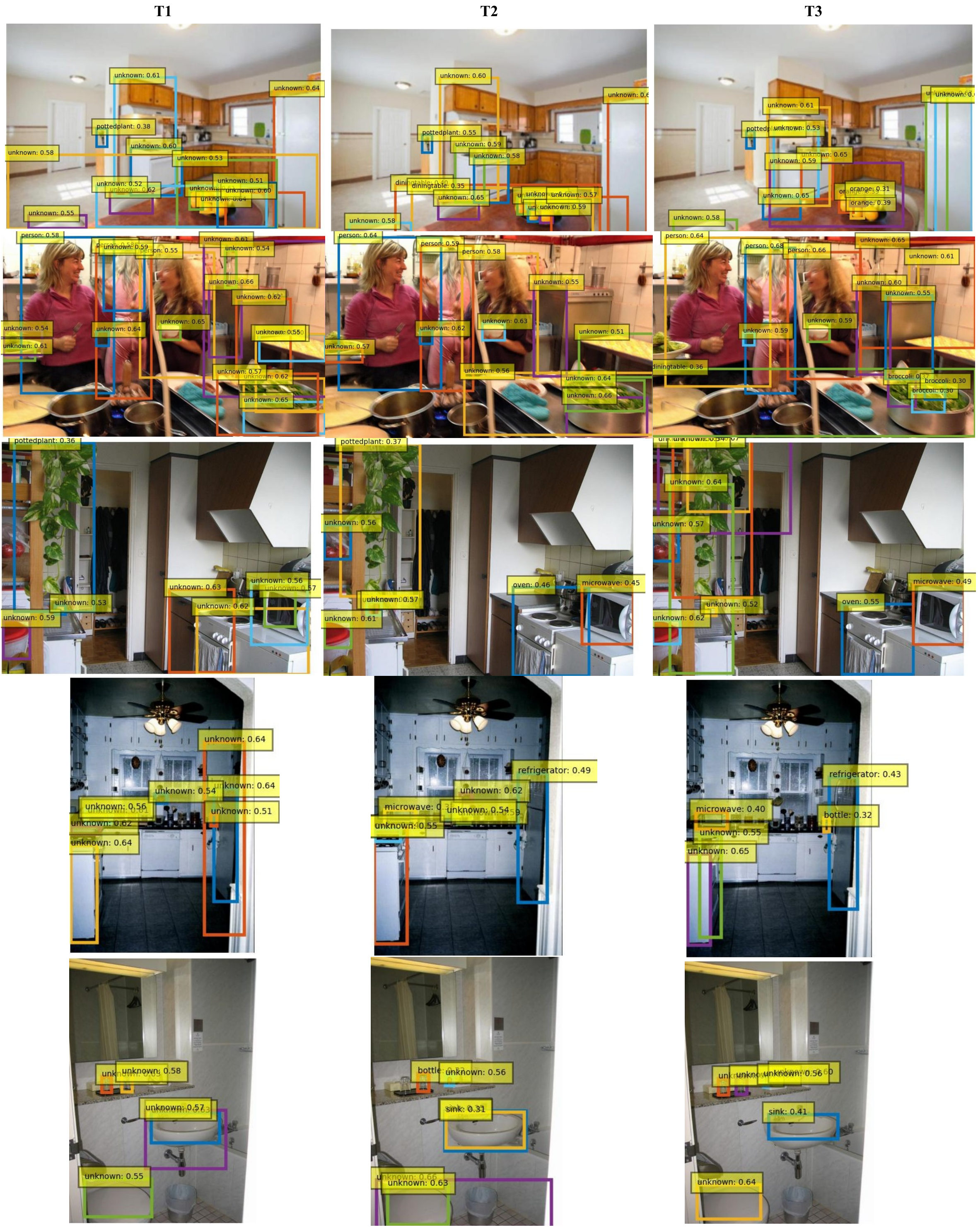}}
    \caption{\textbf{Visualization Results under the Incremental Learning Setting.} 
    We sampled some visualization results of USD in Task 1-3, and some instances were classified from unknown (Task 1) to known classes (Task 2 or Task 3).
    }
    \label{fig:add-vis}
    \vspace{-3mm}
 \end{figure*}

\textbf{Visualization Results under the Incremental Learning Setting:}
As shown in Fig.~\ref{fig:add-vis}, we list some visualization results from Task 1-3.
It is evident that some unknown instances in Task 1 were classified as known classes in Task 2 or Task 3.
For example, \texttt{orange} in the first column, \texttt{broccoli} in the second column, \texttt{microwave} and \texttt{oven} in the third column, \texttt{microwave} and \texttt{refrigerator} in the fourth column, \texttt{sink} in the last column.

\section{Limitations and Social Impacts}

While recent advancements in OWOD have led to significant improvements in unknown recall, the accuracy of unknown object detection remains a critical challenge. 
Because the confused boundary to distinguish objects and backgrounds, labeling unknown objects comprehensively is difficulty and time-consuming.
However, calculating the accuracy metric needs all the unknown objects are fully labeled, so obtaining a test dataset that encompasses all unknown objects with complete labels is a challenging task.
Current OWOD methods tend to prioritize unknown recall, resulting in a high number of false positive instances in unknown predictions. 
This limitation hinders the practical applications of OWOD methods.
For instance, in autonomous driving, the detection of unknown objects or corner cases is crucial to anticipate and handle potential object intrusions, thereby preventing vehicle accidents. 
However, the frequent occurrence of false positive detections in unknown objects leads to frequent warnings, which can become burdensome and impractical.
Therefore, more attention and focus are required to further improve the accuracy of unknown object detection.

Open World Object Detection is a task of great practical significance, particularly in fields like autonomous driving and robotics.
Therefore, OWOD methods will encounter with social impact.
It is crucial to address the challenges related to both recall and accuracy in unknown object detection to ensure the safe and reliable deployment of OWOD methods in real-world scenarios.
The availability of sufficient data and accurate annotation plays a pivotal role in developing models for open environments. 
In this regard, this paper proposes a promising approach that leverages existing Large Vision Models (LVMs), with a specific focus on the Segment Anything Model (SAM), to tackle the annotation dilemma. 
This direction holds great promise for training a robust OWOD method using large-scale data while minimizing the annotation costs. 
\vfill\null
\end{document}